\pdfoutput=1

\documentclass[11pt]{article}

\usepackage[]{naacl2021}

\usepackage{times}
\usepackage{latexsym}
\usepackage{paralist}

\usepackage[T1]{fontenc}

\usepackage[utf8]{inputenc}

\usepackage{microtype}
\usepackage{times}
\usepackage{latexsym}
\usepackage{url}
\usepackage{amsmath}
\usepackage{multirow}
\usepackage{bbm}
\usepackage{graphicx}
\usepackage{amssymb}
\usepackage{booktabs}
\usepackage{adjustbox}
\usepackage{xspace}
\usepackage{algpseudocode}
\usepackage{algorithm}
\usepackage{times}
\usepackage{latexsym}
\usepackage{graphicx}
\usepackage{caption}
\usepackage{subcaption}
\usepackage{color, colortbl}
\usepackage{enumitem}
\usepackage{footnote}
\usepackage{booktabs}
\usepackage{xspace}
\usepackage{makecell}
\makesavenoteenv{tabular}
\makesavenoteenv{table}

\usepackage{array}
\newcommand{\PreserveBackslash}[1]{\let\temp=\\#1\let\\=\temp}
\newcolumntype{C}[1]{>{\PreserveBackslash\centering}p{#1}}
\newcolumntype{R}[1]{>{\PreserveBackslash\raggedleft}p{#1}}
\newcolumntype{L}[1]{>{\PreserveBackslash\raggedright}p{#1}}

\newcommand{\secref}[2][]{Section#1~\ref{sec:#2}}

\newcommand{\tabref}[2][]{Table#1~\ref{tab:#2}}
\newcommand{\figref}[2][]{Figure#1~\ref{fig:#2}}

\title{Discourse Probing of Pretrained Language Models}

\author{Fajri Koto \qquad Jey Han Lau \qquad Timothy Baldwin\\
	School of Computing and Information Systems \\
	The University of Melbourne \\
	\texttt{\small ffajri@student.unimelb.edu.au, jeyhan.lau@gmail.com, 
		tbaldwin@unimelb.edu.au} \\
}

\begin{document}
\maketitle

\begin{abstract}
  Existing work on probing of pretrained language models (LMs) has
  predominantly focused on sentence-level syntactic tasks. In this
  paper, we introduce document-level discourse probing to evaluate the
  ability of pretrained LMs to capture document-level relations. We
  experiment with 7 pretrained LMs, 4 languages, and 7 discourse probing
  tasks, and find BART to be overall the best model at capturing
  discourse --- but only in its encoder, with BERT performing
  surprisingly well as the baseline model. Across the different models,
  there are substantial differences in which layers best capture
  discourse information, and large disparities between
  models.

\end{abstract}

\section{Introduction}

The remarkable development of pretrained language models
\cite{devlin-etal-2019-bert,lewis-etal-2020-bart,lan2020albert}
has raised questions about what precise aspects of language these models
do and do not capture.  Probing tasks offer a means to perform
fine-grained analysis of the capabilities of such models, but most
existing work has focused on sentence-level analysis such as syntax
\cite{hewitt-manning-2019-structural,jawahar-etal-2019-bert,vries2020what},
entities/relations \cite{papanikolaou-etal-2019-deep}, and ontological
knowledge \cite{michael-etal-2020-asking}. Less is known about how well
such models capture broader discourse in documents.

Rhetorical Structure Theory is a framework for capturing how sentences
are connected and describing the overall structure of a document
\cite{mann-thompson-1986-assertions}.  A number of studies have used
pretrained models to classify discourse markers
\cite{sileo-etal-2019-mining} and discourse relations
\cite{nie-etal-2019-dissent,shi-demberg-2019-next}, but few \cite{koto2021top} have
systematically investigated the ability of pretrained models to model
discourse structure.  Furthermore, existing work relating to discourse
probing has typically focused exclusively on the BERT-base model,
leaving open the question of how well these findings generalize to other
models with different pretraining objectives, for different languages,
and different model sizes.

\begin{table}[t]
 	\begin{center}
 		\begin{adjustbox}{max width=1\linewidth}
 			\begin{tabular}{lL{1cm}rrl}
 				\toprule 
 				Model & Type & \#Param & \#Data &Objective \\
 				\midrule 
 				BERT & \multirow{4}{*}{Enc} & 110M & 16GB & MLM+NSP \\
 				RoBERTa & & 110M & 160GB & MLM \\
 				ALBERT & & 12M & 16GB & MLM+SOP \\
 				ELECTRA & & 110M & 16GB & MLM+DISC \\
 				\midrule
 				GPT-2 & Dec & 117M & 40GB & LM \\
 				\midrule
                BART & \multirow{2}{*}{Enc+Dec} & 121M & 160GB & DAE \\
                T5 & & 110M & 750GB & DAE \\
 				\bottomrule
 			\end{tabular}
 		\end{adjustbox}
 	\end{center}
    \caption{Summary of all English pretrained language models used in this
      work. ``MLM'' = masked language model, ``NSP'' = next sentence 
prediction,  ``SOP'' = sentence order prediction, ``LM'' = language model, 
``DISC'' = discriminator, and ``DAE'' = denoising autoencoder.}
	\vspace{-0.3cm}
    \label{tab:models} \end{table}

\begin{table*}[t!]
	\begin{center}
		\begin{adjustbox}{max width=1\linewidth}
			\begin{tabular}{p{4.2cm}p{4.3cm}p{4.3cm}p{4.3cm}p{4cm}}
				\toprule
				\bf Probing Task & \bf English & \bf Chinese & \bf German & \bf Spanish \\
				\midrule
				\makecell[l]{(1) 4-way NSP  \\
					(2) Sentence Ordering
				} & 
				\makecell[l]{XSUM articles \\
					\cite{narayan-etal-2018-dont} \\ 
					Split: 8K/1K/1K
				} &  
				\makecell[l]{Wikipedia (ZH) \\
					Split: 8K/1K/1K
				} &
				\makecell[l]{Wikipedia (DE)\\
					Split: 8K/1K/1K
				} &
				\makecell[l]{Wikipedia (ES)\\
					Split: 8K/1K/1K
				} \\
				
				\midrule
				(3) Discourse Connective & 
				\makecell[l]{Sampled DisSent dataset \\
					\cite{nie-etal-2019-dissent} \\ 
					\#Labels: 15\\
					Split: 10K/1K/1K
				} &  
				\makecell[l]{CDTB \cite{li-etal-2014-building} \\
					\#Labels: 22\\
					Split: 1539/76/168
				} &
				\makecell[l] {Potsdam Commentary \\
					\cite{bourgonje-stede-2020-potsdam}\\
					\#Labels: 15 \\
					Split: 900/148/159
				} &
				\makecell[l]{N/A
				} \\
				
				\midrule		
				\makecell[l]{
					(4) RST Nuclearity\\
					(5) RST Relation\\	
				} & 
				\makecell[l]{RST-DT \\
					\cite{carlson-etal-2001-building} \\ 
					\#Labels (nuc/rel): 3/18 \\
					Split: 16903/1943/2308
				} &  
				\makecell[l]{CDTB \cite{li-etal-2014-building} \\
					\#Labels (nuc/rel): 3/4 \\
					Split: 6159/353/809
				} &
				\makecell[l] {Potsdam Commentary \\
					\cite{bourgonje-stede-2020-potsdam}\\
					\#Labels (nuc/rel): 3/31 \\
					Split: 1892/289/355
				} &
				\makecell[l]{RST-Spanish Treebank\\
					\cite{cunha2011on}\\
					\#Labels (nuc/rel): 3/29 \\
					Split: 2042/307/421
				} \\
				
				\midrule
				(6) RST EDU Segmentation & 
				\makecell[l]{RST-DT \\
					\cite{carlson-etal-2001-building} \\ 
					Split: 312/35/38 docs
				} &  
				\makecell[l]{CDTB \cite{li-etal-2014-building} \\
					Split: 2135/105/241 p'graphs
				} &
				\makecell[l] {Potsdam Commentary \\
					\cite{bourgonje-stede-2020-potsdam}\\
					Split: 131/20/25 docs
				} &
				\makecell[l]{RST-Spanish Treebank\\
					\cite{cunha2011on}\\
					Split: 200/34/30 docs
				} \\
				
				\midrule
				(7) Cloze Story Test & 
				\makecell[l]{
					\cite{mostafazadeh-etal-2016-corpus}\\ 
					Split: 1683/188/1871
				} &  
				N/A & N/A & N/A \\
				
				\bottomrule
			\end{tabular}
		\end{adjustbox}
	\end{center}
    \caption{A summary of probing tasks and datasets for each of the
      four languages. ``Split'' indicates 
    the number of  train/development/test instances.}
	\label{tab:data}
\end{table*}

Our research question in this paper is: \textit{How much discourse
  structure do layers of different pretrained language models capture,
  and do the findings generalize across languages?}
 
There are two contemporaneous related studies that have examined discourse 
modelling in pretrained language models.  
\citet{upadhye-etal-2020-predicting} analyzed how well two pretrained
models capture referential biases of different classes of English verbs.  
\citet{zhu-etal-2020-examining} applied the model of  
\citet{feng-hirst-2014-linear} to parse IMDB documents 
\cite{maas-etal-2011-learning} into discourse trees. Using this 
(potentially noisy) data,
probing tasks were conducted by mapping attention layers into single
vectors of document-level rhetorical features. 
These features, however, are unlikely to capture all the intricacies of
inter-sentential abstraction as their input is formed based on discourse
relations\footnote{For example, they only consider discourse relation
  labels and ignore nuclearity.} and aggregate statistics on
the distribution of discourse units.

To summarize, we introduce 7 discourse-related probing tasks, which we
use to analyze 7 pretrained language models over 4 languages: English,
Mandarin Chinese, German, and Spanish.  Code and public-domain data
associated with this research is available at
\url{https://github.com/fajri91/discourse_probing}.




\section{Pretrained Language Models}

We outline the 7 pretrained models in \tabref{models}.  They comprise 4
encoder-only models: BERT \cite{devlin-etal-2019-bert}, RoBERTa
\cite{liu2019roberta}, ALBERT \cite{lan2020albert}, and ELECTRA
\cite{clark2020electra}; 1 decoder-only model: GPT-2
\cite{radford2019language}; and 2 encoder--decoder models: BART
\cite{lewis-etal-2020-bart} and T5 \cite{raffel2019exploring}. To reduce
the confound of model size, we use pretrained models of similar size
($\sim$110m model parameters), with the exception of ALBERT
which is designed to be lighter weight. All models have 12 transformer
layers in total; for BART and T5, this means their encoder and decoder
have 6 layers each.  Further details of the models are provided in the Supplementary Material.

\begin{figure}[t]
	\centering
	\includegraphics[width=3in]{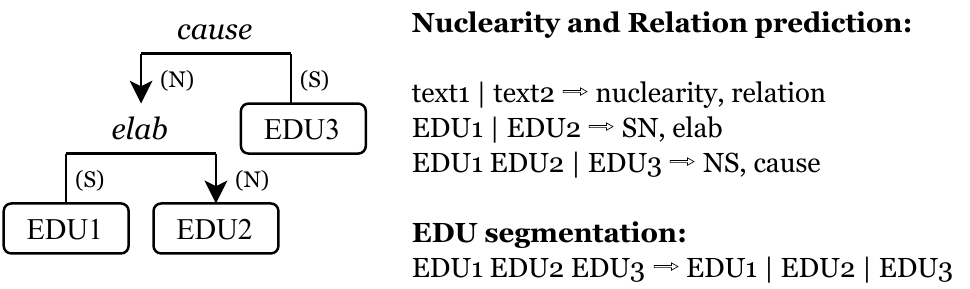}
    \caption{\label{fig:rst} Illustration of the RST discourse probing
      tasks (Tasks 4--6).}
\end{figure}

\begin{figure*}[t!]
	\vspace{-0.2cm}
	\centering
	\includegraphics[width=6.1in]{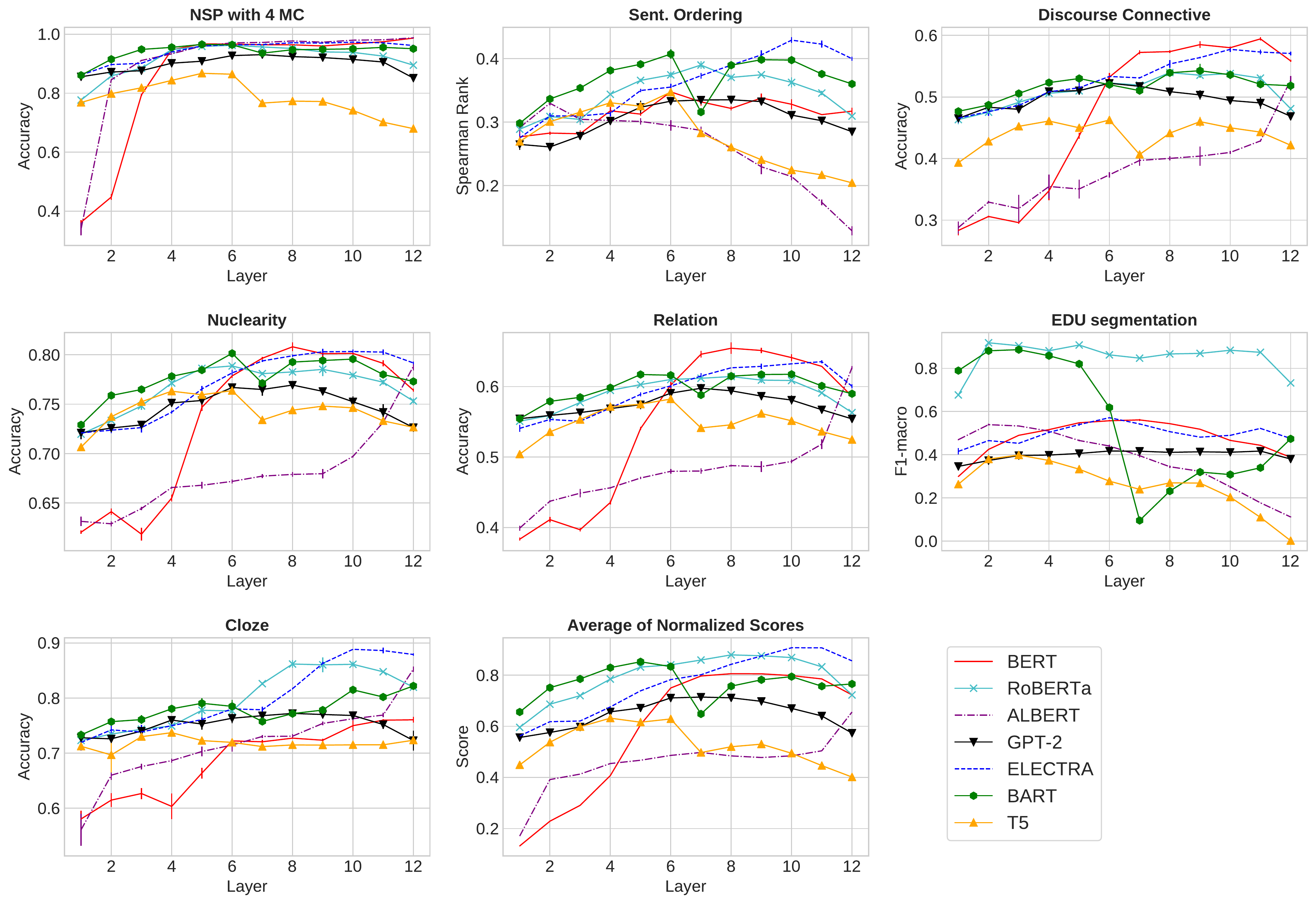}
        \caption{Probing task performance on English for each of the
          seven tasks, plus the average across all tasks. For BART and
          T5, layers 7--12 are the decoder layers. All results are
          averaged over three runs, and the vertical line for each data point
          denotes the standard deviation (noting that most results have low
          s.d., meaning the bar is often not visible).}
	\label{fig:result_en}
\end{figure*}

\section{Probing Tasks for Discourse Coherence}
\label{sec:tasks}

We experiment with a total of seven probing tasks, as detailed
below. Tasks 4--6 are component tasks of discourse parsing based on
rhetorical structure theory (RST;
\citet{mann-thompson-1986-assertions}). In an RST discourse tree, EDUs are typically
clauses or sentences, and are hierarchically connected with discourse 
labels denoting: (1) nuclearity = nucleus (N) vs.\ satellite 
(S);\footnote{The satellite is a supporting EDU for the nucleus.} 
and (2) discourse relations (e.g.\ \textit{elaborate}).  An example of a
binarized RST discourse tree is given in \figref{rst}.

\begin{enumerate}
\item \textbf{Next sentence prediction.} Similar to the next sentence
  prediction (NSP) objective in BERT pretraining, but here we frame it
  as a 4-way classification task, with one positive and 3 negative
  candidates for the next sentence.  The preceding context takes the
  form of between 2 and 8 sentences, but the candidates are always
  single sentences.

\item \textbf{Sentence ordering.} We shuffle 3--7 sentences and attempt
  to reproduce the original order. This task is based on
  \citet{barzilay-lapata-2008-modeling} and \citet{koto2020indolem}, and
  is assessed based on rank correlation relative to the original order.

\item \textbf{Discourse connective prediction.} Given two
  sentences/clauses, the task is to identify an appropriate discourse
  marker, such as \textit{while}, \textit{or}, or \textit{although}
  \cite{nie-etal-2019-dissent}, representing the conceptual relation
  between the sentences/clauses.

\item \textbf{RST nuclearity prediction.} For a given ordered pairing of
  (potentially complex) EDUs which are connected by an unspecified
  relation, predict the nucleus/satellite status of each (see \figref{rst}).
  
\item \textbf{RST relation prediction.} For a given ordered pairing of
  (potentially complex) EDUs which are connected by an unspecified
  relation, predict the relation that holds between them (see
  \figref{rst}).

\item \textbf{RST elementary discourse unit (EDU) segmentation.}  Chunk a
  concatenated sequence of EDUs into its component EDUs.

\item \textbf{Cloze story test.} Given a 4-sentence story context, pick
  the best ending from two possible options
  \cite{mostafazadeh-etal-2016-corpus,sharma-etal-2018-tackling}. This
  task is harder than NSP, as it requires an understanding of
  commonsense and storytelling \cite{chaturvedi-etal-2017-story,liu-etal-2018-narrative}.
\end{enumerate}

\section{Experimental Setup}

We summarize all data (sources, number of labels, and data split) in
\tabref{data}. This includes English, Chinese, German, and Spanish for
each probing task. For NSP and sentence ordering, we generate data from
news articles and Wikipedia. For the RST tasks, we use discourse treebanks
for each of the four languages.


We formulate all probing tasks except sentence ordering and EDU
segmentation as a classification problem, and evaluate using
accuracy. During fine-tuning, we add an MLP layer on top of the
pretrained model for classification, and only update the MLP parameters
(all other layers are frozen).  We use the [CLS] embedding for BERT and
ALBERT following standard practice,
while for other models we perform average pooling to obtain a vector for
each sentence, and concatenate them as the input to the
MLP.\footnote{BERT and ALBERT performance with average pooling is in
  included in the Appendix.}


For sentence ordering, we follow \citet{koto2020indolem} and frame it as 
a sentence-level sequence labelling task, where the goal is to estimate 
$P(r|s)$, where $r$ is the rank position and $s$ the sentence. The task 
has 7 classes, as we have 3--7 sentences (see \secref{tasks}). At test time,
we choose the label sequence that maximizes the sequence probability.
Sentence embeddings are obtained by average pooling. The EDU
segmentation task is also framed as a binary sequence labelling task
(segment boundary or not) at the (sub)word level. 
We use Spearman rank correlation and macro-averaged F1 score to evaluate sentence ordering and EDU segmentation, respectively.



We use a learning rate $1e-3$, warm-up of 10\% of total steps, and the
development set for early stopping in all experiments. All presented
results are averaged over three runs.\footnote{More details of the
  training configuration are given in the Appendix.}

\begin{figure*}
	\vspace{-0.25cm}
	\centering
	\includegraphics[width=6.2in]{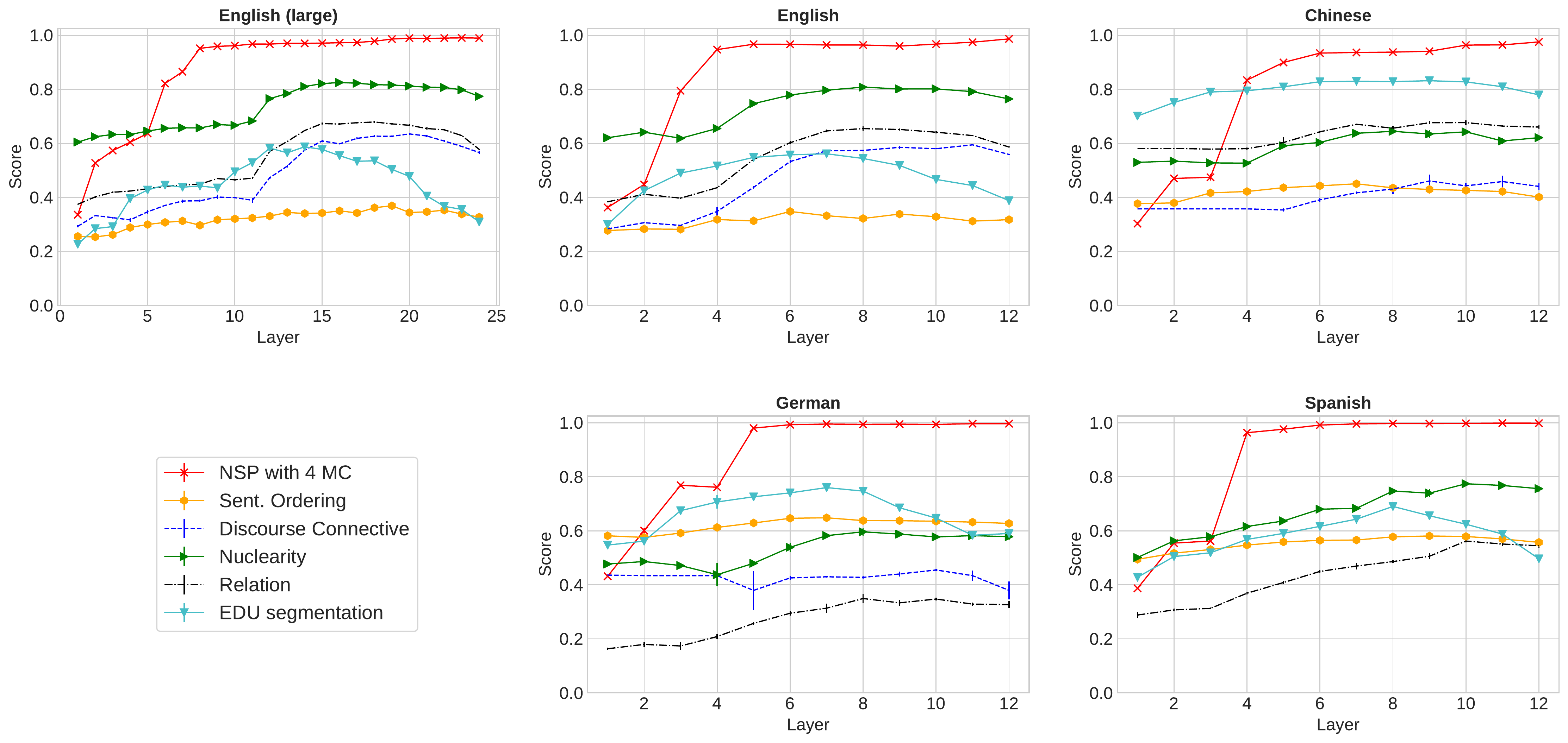}
        \caption{Discourse performance of BERT across different
          languages. All results are averaged over three runs, and a
          vertical line is used to denote the standard deviation for
          each data point (most of which are not visible, due to the low standard deviation).}
	\label{fig:result_other}
\end{figure*}

\section{Results and Analysis}

In \figref{result_en}, we present the probing task performance on
English for all models based on a representation generated from each of
the 12 layers of the model. First, we observe that most performance fluctuates (non-monotonic) across layers except for some models in the NSP task and some ALBERT results in the other probing tasks.
We also found that most models
except ALBERT tend to have a very low standard deviation based on three
runs with different random seeds.

\begin{figure}
	\centering
    \includegraphics[width=3.1in]{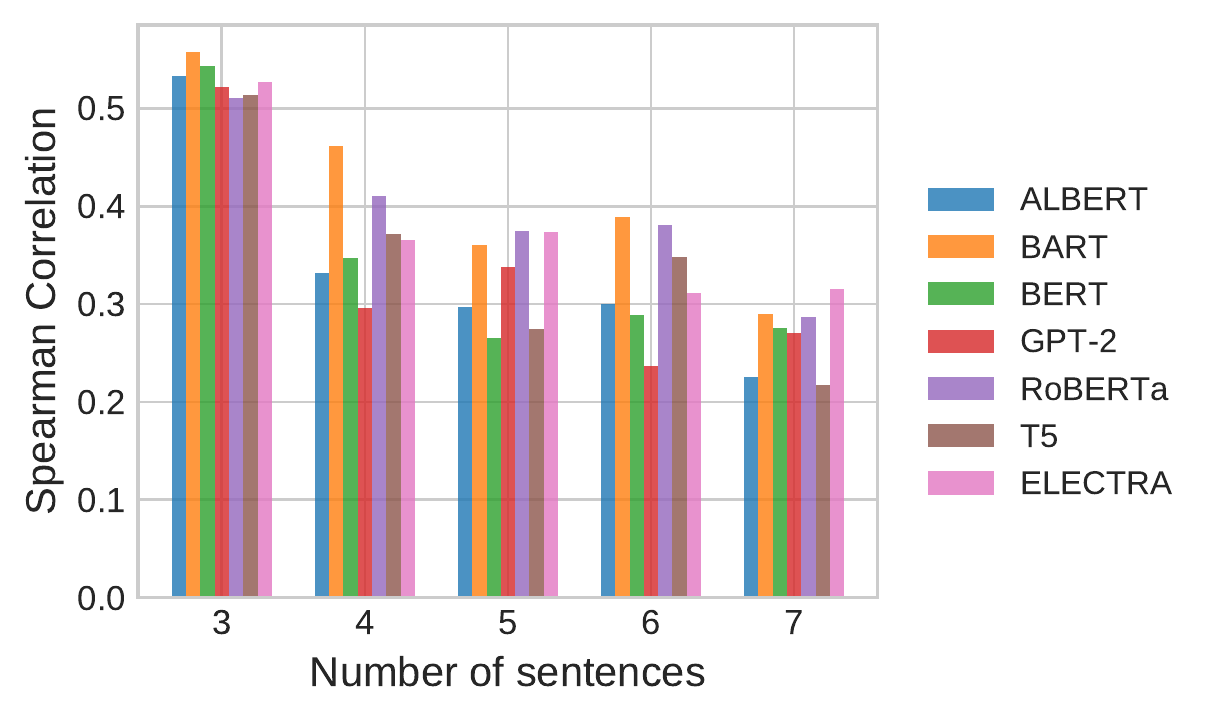}
	\caption{Sentence ordering task breakdown based on the best layer of each model.}
	\label{fig:analysis}
	\vspace{-0.3cm}
\end{figure}

We discover that all models except T5
and early layers of BERT and ALBERT perform well over the NSP task, with
accuracy $\geq$ 0.8, implying it is a simple task. However, they all
struggle at sentence ordering (topping out at $\rho \sim 0.4$),
suggesting that they are ineffective at modelling discourse over multiple
sentences;  this is borne out in \figref{analysis}, where
performance degrades as the number of sentences to re-order increases.



Interestingly, for Discourse Connectives, RST Nuclearity, and RST
Relation Prediction, the models produce similar patterns, even though
the discourse connective data is derived from a different dataset and
theoretically divorced from RST. BART outperforms most other models in
layers 1--6 for these tasks (a similar observation is found for NSP and
Sentence Ordering) with BERT and ALBERT struggling particularly in the
earlier layers. For EDU segmentation, RoBERTa and again the first few
layers of BART perform best. For the Cloze Story Test, all models seem
to improve as we go deeper into the layers, suggesting that high-level
story understanding is captured deeper in the models.

We summarize the overall performance by calculating the averaged
normalized scores in the last plot in \figref{result_en}.\footnote{Given
  a task, we perform min--max normalization for all model-layer scores
  (7$\times$12 scores in total), and then compute the average over all
  tasks for each model's layer.}  RoBERTa and BART appear to be the best
overall models at capturing discourse information, but only in the
\textit{encoder layers} (the first 6 layers) for BART. We hypothesize
that the BART decoder focuses on sequence generation, and as such is
less adept at language understanding.  This is supported by a similar
trend for T5, also a denoising autoencoder. BERT does surprisingly well
(given that it's the baseline model), but mostly in the deeper layers
(7--10), while ELECTRA performs best at the three last layers.




In terms of the influence of training data, we see mixed results. BART
and RoBERTa are the two best models, and both are trained with more data
than most models (an order of magnitude more; see \tabref{models}).  But
T5 (and to a certain extent GPT-2) are also trained with more data (in
fact T5 has the most training data), but their discourse modelling
performance is underwhelming. In terms of training objectives, it
appears that a pure decoder with an LM objective (GPT-2) is less
effective at capturing discourse structure.  ALBERT, the smallest model
(an order of magnitude less parameters than most), performs surprisingly
well (with high standard deviation), but only at its last layer,
suggesting that discourse knowledge is concentrated deep inside the
model.


Lastly, we explore whether these trends hold if we use a larger model
(BERT-base vs.\ BERT-large) and for different languages (again based on
monolingual BERT models for the respective languages). Results are
presented in \figref{result_other}. For model size (``English (large)''
vs.\ ``English''), the overall pattern is remarkably similar, with a
slight uplift in absolute results with the larger model.  Between the 4
different languages (English, Chinese, German, and Spanish), performance
varies for all tasks except for NSP (e.g.\ EDU segmentation appears to
be easiest in Chinese, and relation prediction is the hardest in
German), but the \textit{shape} of the lines is largely the same,
indicating the optimal layers for a particular task are consistent
across languages.


\section{Conclusion}

We perform probing on 7 pretrained language models across 4 languages to
investigate what discourse effects they capture.  We find that BART's
encoder and RoBERTa perform best, while pure language models (GPT-2)
struggle. Interestingly, we see a consistent pattern across different
languages and model sizes, suggesting that the trends we found are
robust across these dimensions.


%

\section*{Acknowledgements}
We are grateful to the anonymous reviewers for their helpful feedback
and suggestions. The first author is supported by the Australia Awards
Scholarship (AAS), funded by the Department of Foreign Affairs and Trade
(DFAT), Australia.  This research was undertaken using the LIEF
HPC-GPGPU Facility hosted at The University of Melbourne. This facility
was established with the assistance of LIEF Grant LE170100200.

\bibliography{anthology,custom}
\bibliographystyle{acl_natbib}

\clearpage
\appendix

\section{Pretrained Language Models}
\label{sec:models}
	
The pretrained models are sourced from Huggingface
(\url{https://huggingface.co/}), as detailed in \tabref[s]{en_model}
and \ref{tab:non_en_model}.

\begin{table}[h]
	\begin{center}
		\begin{adjustbox}{max width=1\linewidth}
			\begin{tabular}{ll}
				\toprule
				\textbf{Model} & \bf Huggingface model \\
				\midrule
				BERT & \texttt{bert-base-uncased} \\
				BERT (large) & \texttt{bert-large-uncased} \\
				RoBERTa & \texttt{roberta-base} \\
				ALBERT & \texttt{albert-base-v2} \\
				ELECTRA & \texttt{electra-base-discriminator} \\
				\midrule
				GPT-2 & \texttt{gpt2} \\
				\midrule
				BART & \texttt{bart-base} \\
				T5 & \texttt{t5-small} \\
				\bottomrule
			\end{tabular}
		\end{adjustbox}
	\end{center}
	\caption{List of English pretrained language models}
	\label{tab:en_model}
\end{table}

\begin{table}[h]
	\begin{center}
		\begin{adjustbox}{max width=1\linewidth}
			\begin{tabular}{ll}
				\toprule
				\textbf{Language} & \bf Huggingface model \\
				\midrule
				Chinese & \texttt{bert-base-chinese} \\
				German & \texttt{bert-base-german-dbmdz-uncased} \\
				Spanish & \texttt{bert-base-spanish-wwm-uncased} \\
				\bottomrule
			\end{tabular}
		\end{adjustbox}
	\end{center}
	\caption{List of non-English BERT models.}
	\label{tab:non_en_model}
\end{table}

\section{Data Construction, Examples, and Training Configuration}
\label{sec:construction}

\subsection{Next Sentence Prediction}

\begin{table}[t!]
	\begin{center}
		\begin{adjustbox}{max width=0.6\linewidth}
			\begin{tabular}{rr}
				\toprule
				\textbf{\#Sentence (context)} & \bf Total \\
				\midrule
				2 & 2500 \\
				4 & 2500 \\
				6 & 2500 \\
				8 & 2500 \\
				\midrule
				Total & 10000 \\
				\bottomrule
			\end{tabular}
		\end{adjustbox}
	\end{center}
	\caption{NSP data based on the number of sentences.}
	\label{tab:b1_data}
\end{table}

We use spaCy (\url{https://spacy.io/}) to perform sentence tokenization, and ensure
that the distractor options in the training set do not overlap with the
test set. For all languages and models, the training configurations are
similar: the maximum tokens in the context and the next sentence are 450
and 50, respectively. If the token lengths are more than this, we
truncate the context from the beginning of the sequence, and truncate
the next sentence at the end of the sequence. We concatenate context
with each option, and perform binary classification.

Other training configuration details: learning rate = 1e-3, Adam epsilon
=1e-8, maximum gradient norm = 1.0, maximum epochs = 20, warmup = 10\%
of the training steps, and patience for early stopping = 5 epochs.

\begin{table}[t]
	\begin{center}
		\begin{adjustbox}{max width=0.8\linewidth}
			\begin{tabular}{p{7cm}}
				\toprule
				\textbf{Context} \\
				\midrule
				\textbf{s1:} The Eastern Star, mostly carrying elderly tourists, capsized on 1 June near Jianli in Hubei province.\\
				\textbf{s2:} Just 14 of the 456 passengers and crew are known to have survived. \\
				\midrule
				\textbf{Options}\\
				\midrule
				\textbf{0:} The channel recently said its signal was carried by 22 satellites \\
				\textbf{0:} That step has become a huge challenge for opposition candidates \\
				\textbf{0:} Six men were convicted and then acquitted of the atrocity and no-one has since been convicted of involvement in the bombing \\
				\textbf{1:} A search is continuing for eight people who remain missing.\\
				\bottomrule
			\end{tabular}
		\end{adjustbox}
	\end{center}
	\caption{Example of English NSP data with 2-sentence context. 1 indicates the correct next sentence.}
	\label{tab:b1_ex}
\end{table}

\subsection{Sentence Ordering}

\begin{table}[t!]
	\begin{center}
		\begin{adjustbox}{max width=0.4\linewidth}
			\begin{tabular}{rr}
				\toprule
				\textbf{\#Sentence} & \bf Total \\
				\midrule
				3 & 2000 \\
				4 & 2000 \\
				5 & 2000 \\
				6 & 2000 \\
				7 & 2000 \\
				\midrule
				Total & 10000 \\
				\bottomrule
			\end{tabular}
		\end{adjustbox}
	\end{center}
	\caption{Sentence ordering data based on number of sentence.}
	\label{tab:b2_data}
\end{table}

\begin{table}[t!]
	\begin{center}
		\begin{adjustbox}{max width=0.8\linewidth}
			\begin{tabular}{p{7cm}}
				\toprule
				\textbf{Context} \\
				\midrule
				\textbf{s0:} West Mercia Police said the police do not encourage members of the public to pursue their own investigations.\\
				\textbf{s1:} David John Poole, from Hereford, poses online as a 14-year-old girl and says he has been sent hundreds of explicit messages. \\
				\textbf{s2:} He says his work has led to two arrests in four weeks. \\
				\midrule
				\textbf{Correct order:} 2--0--1\\
				
				\bottomrule
			\end{tabular}
		\end{adjustbox}
	\end{center}
	\caption{Example of English sentence ordering data}
	\label{tab:b2_ex}
\end{table}

In generating sentence ordering data, we once again use spaCy
(\url{https://spacy.io/}) to perform sentence tokenization. For all
languages and models, the training configurations are similar, with the
maximum tokens in each sentence = 50, learning rate = 1e-3, Adam epsilon
= 1e-8, maximum gradient norm = 1.0, training epochs = 20, warmup = 10\%
of the training steps, and patience for early stopping = 10 epochs. 

\subsection{Discourse Connective Prediction}

\begin{figure}[t!]
	\centering
	\includegraphics[width=2.8in]{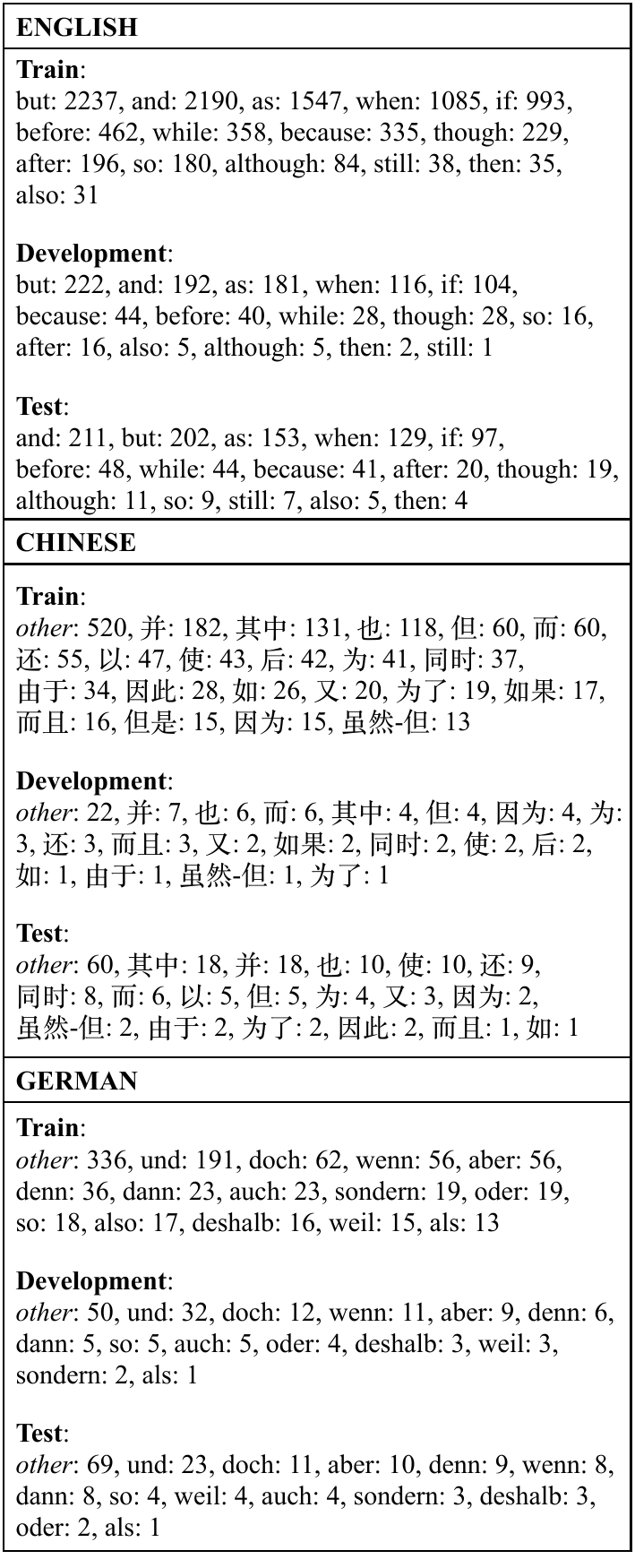}
	\caption{Discourse connective word distribution.}
	\label{fig:data_b3}
\end{figure}

As our Chinese and German data is extracted from discourse
treebanks, the number of distinct connective words varies. For instance,
in the Chinese discourse treebank, we find 246 unique connective words. To simplify this, we set the connective word to \textit{OTHER} if its word frequency is less than 12.

For all languages and models, the training configurations are: maximum
token length of each sentence = 50, learning rate = 1e-3, Adam epsilon =
1e-8, maximum gradient norm = 1.0, maximum epochs = 20, warmup = 10\% of
the training steps, and patience for early stopping = 10 epochs. 

\begin{figure}[t!]
	\centering
	\includegraphics[width=2.8in]{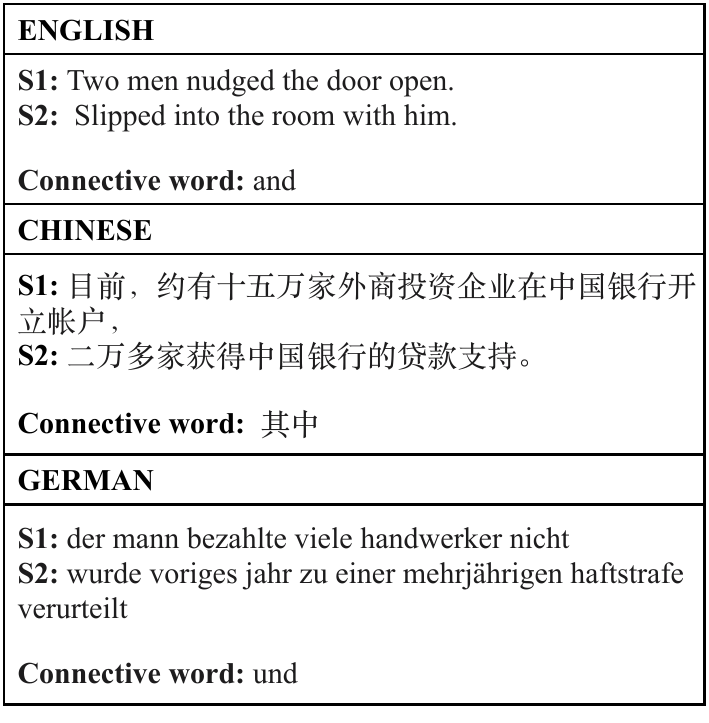}
	\caption{Discourse connective: data examples}
	\label{fig:data_exb3}
\end{figure}

\begin{figure}[t!]
	\centering
	\includegraphics[width=2.8in]{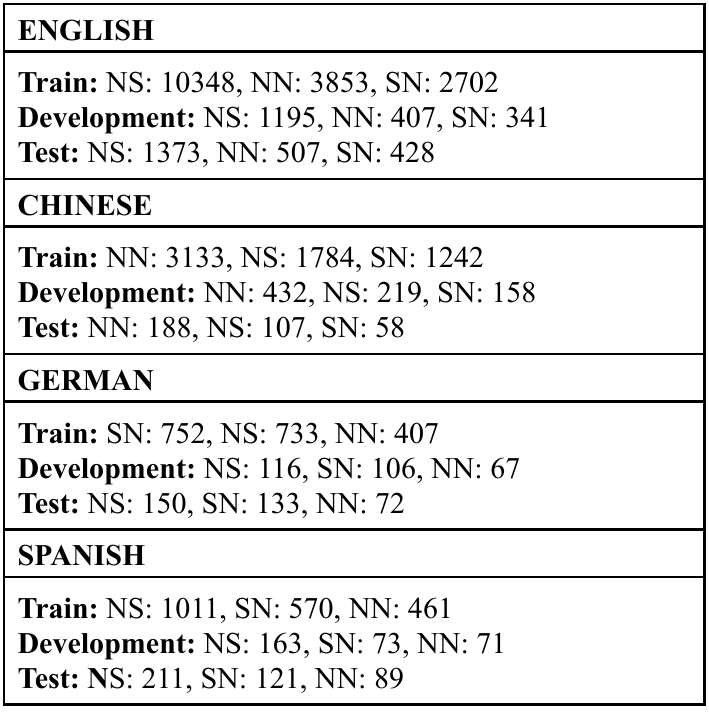}
	\caption{Nuclearity label distribution.}
	\label{fig:data_b4_nuc}
\end{figure}

\begin{figure}[t!]
	\centering
	\includegraphics[width=2.8in]{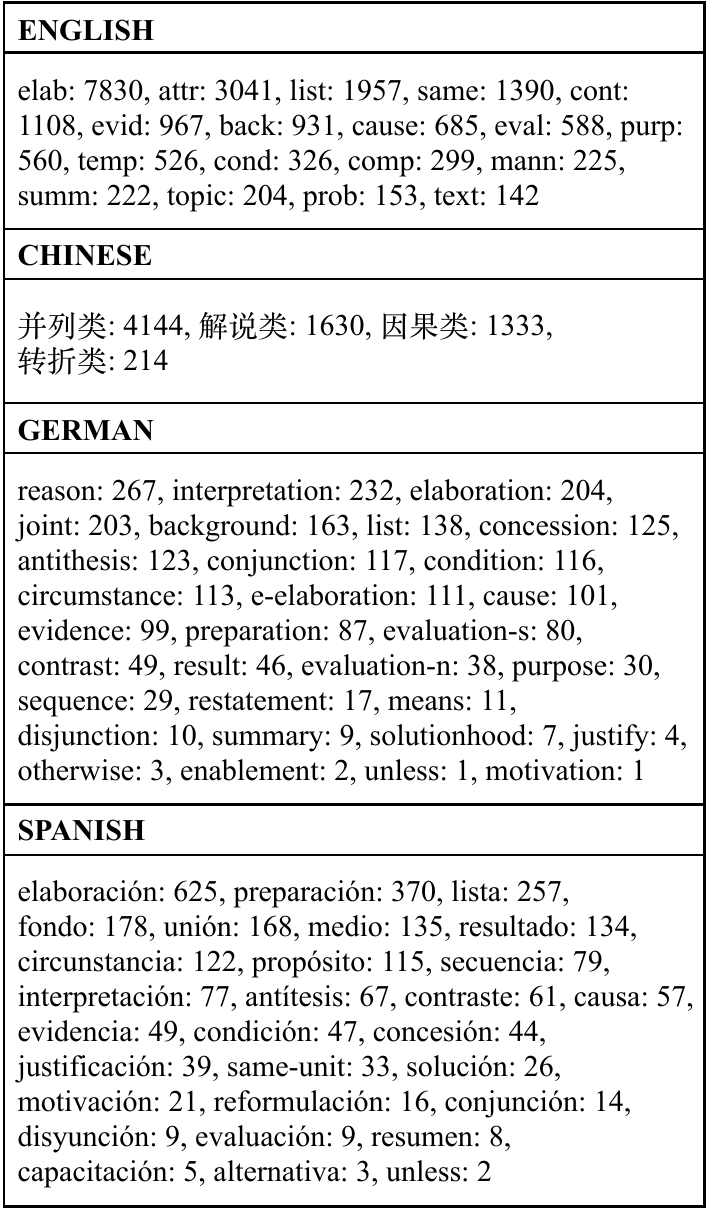}
	\caption{Relation label distribution.}
	\label{fig:data_b4_rel}
\end{figure}

\subsection{RST-related Tasks}

In \figref[s]{data_b4_nuc} and  \ref{fig:data_b4_rel}, we present the
distribution of the nuclearity and relation labels for the 4 different
discourse treebanks. The English treebank is significantly larger, with
a strong preference for the NS (nuclear--satellite) relationship. Unlike
other languages, the proportion of NN (nuclear--nuclear) relationships
in the Chinese discourse treebank (CDTB) is the highest. We also notice
that the relation label set in CDTB is the simplest, with only 4 labels.

Most of the training details for nuclearity and relation prediction are
the same as for the NSP task, except we set the maximum token length of
each sentence to 250. Particularly for EDU segmentation, we set the
maximum token length in a document to 512.

\subsection{Cloze Story Test}

As discussed in \tabref{data}, we use cloze story test version-1
\cite{mostafazadeh-etal-2016-corpus}. Although version-2
\cite{sharma-etal-2018-tackling} is better in terms of story biases, the
gold labels for the test set are not publicly available, which limited
our ability to explore different layers of a broad range of pretrained
language models (due to rate limiting of test evaluation). 

For the data split, we followed previous work
\cite{liu-etal-2018-narrative} in splitting the development set into a
training and validation set. We perform binary classification similar to
the NSP task, by first merging all 4-sentence stories into a single text
(context). We limit the context to a maximum of 450 tokens, and each
candidate sentence (as the story ending) is limited to 50 tokens. Other
training details are the same as for the NSP task.

\clearpage
\onecolumn
\section{Full Experimental Results}
\label{sec:complete}

	\begin{center}
	\begin{adjustbox}{width=0.75\linewidth}
		\begin{tabular}{R{0.7cm}R{1.5cm}R{1.5cm}R{1.5cm}R{1.5cm}R{1.5cm}R{1.5cm}R{1.5cm}}
			\toprule
			\bf Layer &\bf  NSP & \bf Sent. Ord. & \bf Discourse Conn. & \bf Nuclearity & \bf Relation & \bf EDU segment. & \bf Cloze ST. \\
				\midrule
				\multicolumn{8}{l}{BERT (English); std = 0.00 -- 0.02} \\
				\midrule
				1 & 0.36 & 0.28 & 0.28 & 0.62 & 0.38 & 0.30 & 0.58 \\
				2 & 0.45 & 0.28 & 0.31 & 0.64 & 0.41 & 0.42 & 0.61 \\
				3 & 0.79 & 0.28 & 0.30 & 0.62 & 0.40 & 0.49 & 0.63 \\
				4 & 0.95 & 0.32 & 0.35 & 0.65 & 0.44 & 0.52 & 0.60 \\
				5 & 0.97 & 0.31 & 0.44 & 0.75 & 0.54 & 0.55 & 0.66 \\
				6 & 0.97 & \bf 0.35 & 0.53 & 0.78 & 0.60 & \bf 0.56 & 0.72 \\
				7 & 0.96 & 0.33 & 0.57 & 0.80 & \bf 0.65 & \bf 0.56 & 0.72 \\
				8 & 0.96 & 0.32 & 0.57 & \bf 0.81 & \bf 0.65 & 0.54 & 0.73 \\
				9 & 0.96 & 0.34 & \bf 0.59 & 0.80 & \bf 0.65 & 0.52 & 0.72 \\
				10 & 0.97 & 0.33 & 0.58 & 0.80 & 0.64 & 0.47 & 0.75 \\
				11 & 0.97 & 0.31 & \bf 0.59 & 0.79 & 0.63 & 0.44 & \bf 0.76 \\
				12 & \bf 0.99 & 0.32 & 0.56 & 0.76 & 0.59 & 0.39 & \bf 0.76 \\
			
				\midrule
				\multicolumn{8}{l}{RoBERTa (English); std = 0.00 -- 0.02} \\
				\midrule
				1 & 0.78 & 0.29 & 0.46 & 0.72 & 0.55 & 0.68 & 0.72 \\
				2 & 0.86 & 0.31 & 0.48 & 0.73 & 0.56 & \bf 0.92 & 0.73 \\
				3 & 0.88 & 0.30 & 0.49 & 0.75 & 0.58 & 0.90 & 0.74 \\
				4 & 0.95 & 0.34 & 0.51 & 0.77 & 0.59 & 0.88 & 0.75 \\
				5 & \bf 0.96 & 0.37 & 0.51 & \bf 0.79 & 0.60 & 0.91 & 0.78 \\
				6 & \bf 0.96 & 0.37 & 0.52 & \bf 0.79 & \bf 0.61 & 0.86 & 0.78 \\
				7 & \bf 0.96 & \bf 0.39 & 0.52 & 0.78 & \bf 0.61 & 0.85 & 0.83 \\
				8 & 0.95 & 0.37 & \bf 0.54 & 0.78 & \bf 0.61 & 0.87 & \bf 0.86 \\
				9 & 0.94 & 0.37 & \bf 0.54 & \bf 0.79 & \bf 0.61 & 0.87 & \bf 0.86 \\
				10 & 0.94 & 0.36 & \bf 0.54 & 0.78 & \bf 0.61 & 0.88 & \bf 0.86 \\
				11 & 0.93 & 0.35 & 0.53 & 0.77 & 0.59 & 0.87 & 0.85 \\
				12 & 0.90 & 0.31 & 0.48 & 0.75 & 0.56 & 0.73 & 0.82 \\
				
				\midrule
				\multicolumn{8}{l}{ALBERT (English); std = 0.00 -- 0.03} \\
				\midrule
				1 & 0.34 & 0.29 & 0.29 & 0.63 & 0.40 & 0.47 & 0.56 \\
				2 & 0.85 & \bf 0.33 & 0.33 & 0.63 & 0.44 & \bf 0.54 & 0.66 \\
				3 & 0.91 & 0.30 & 0.32 & 0.64 & 0.45 & 0.53 & 0.68 \\
				4 & 0.93 & 0.30 & 0.35 & 0.67 & 0.46 & 0.51 & 0.69 \\
				5 & 0.96 & 0.30 & 0.35 & 0.67 & 0.47 & 0.47 & 0.70 \\
				6 & 0.97 & 0.29 & 0.37 & 0.67 & 0.48 & 0.44 & 0.71 \\
				7 & 0.97 & 0.29 & 0.40 & 0.68 & 0.48 & 0.40 & 0.73 \\
				8 & 0.98 & 0.26 & 0.40 & 0.68 & 0.49 & 0.34 & 0.73 \\
				9 & 0.97 & 0.23 & 0.40 & 0.68 & 0.49 & 0.32 & 0.75 \\
				10 & 0.98 & 0.21 & 0.41 & 0.70 & 0.49 & 0.25 & 0.76 \\
				11 & 0.98 & 0.17 & 0.43 & 0.73 & 0.52 & 0.18 & 0.77 \\
				12 & \bf 0.99 & 0.13 & \bf 0.53 & \bf 0.79 & \bf 0.63 & 0.11 & \bf 0.85 \\

				\midrule
				\multicolumn{8}{l}{ELECTRA (English); std = 0.00 -- 0.02} \\
				\midrule
				1 & 0.86 & 0.27 & 0.47 & 0.72 & 0.54 & 0.42 & 0.72 \\
				2 & 0.90 & 0.31 & 0.48 & 0.72 & 0.55 & 0.47 & 0.74 \\
				3 & 0.90 & 0.31 & 0.49 & 0.73 & 0.55 & 0.45 & 0.74 \\
				4 & 0.94 & 0.31 & 0.51 & 0.74 & 0.57 & 0.50 & 0.75 \\
				5 & 0.96 & 0.35 & 0.52 & 0.77 & 0.59 & 0.54 & 0.76 \\
				6 & 0.96 & 0.36 & 0.53 & 0.78 & 0.60 & \bf 0.57 & 0.78 \\
				7 & 0.96 & 0.37 & 0.53 & 0.79 & 0.61 & 0.54 & 0.78 \\
				8 & \bf 0.97 & 0.39 & 0.55 & \bf 0.80 & 0.63 & 0.51 & 0.82 \\
				9 & \bf 0.97 & 0.41 & 0.56 & \bf 0.80 & 0.63 & 0.48 & 0.86 \\
				10 & \bf 0.97 & \bf 0.43 & \bf 0.58 & \bf 0.80 & 0.63 & 0.49 & \bf 0.89 \\
				11 & \bf 0.97 & 0.42 & 0.57 & \bf 0.80 & \bf 0.64 & 0.52 & \bf 0.89 \\
				12 & 0.96 & 0.40 & 0.57 & 0.79 & 0.60 & 0.48 & 0.88 \\

				\bottomrule		
			\end{tabular}
		\end{adjustbox}
	\captionof{table}{Full results for BERT, RoBERTa, ALBERT, and
          ELECTRA over English.}
	\label{tab:result_p1}
	\end{center}

	\vspace{1cm}

	\begin{center}
		\begin{adjustbox}{width=0.75\linewidth}
			\begin{tabular}{R{0.7cm}R{1.5cm}R{1.5cm}R{1.5cm}R{1.5cm}R{1.5cm}R{1.5cm}R{1.5cm}}
				\toprule
				\bf Layer &\bf  NSP & \bf Sent. Ord. & \bf Discourse Conn. & \bf Nuclearity & \bf Relation & \bf EDU segment. & \bf Cloze ST. \\
				
				\midrule
				\multicolumn{8}{l}{GPT-2 (English); std = 0.00 -- 0.02} \\
				\midrule
				1 & 0.86 & 0.26 & 0.47 & 0.72 & 0.55 & 0.35 & 0.73 \\
				2 & 0.87 & 0.26 & 0.48 & 0.73 & 0.56 & 0.37 & 0.73 \\
				3 & 0.88 & 0.28 & 0.48 & 0.73 & 0.56 & 0.40 & 0.74 \\
				4 & 0.90 & 0.30 & 0.51 & 0.75 & 0.57 & 0.40 & 0.76 \\
				5 & 0.91 & 0.32 & 0.51 & 0.75 & 0.57 & 0.41 & 0.75 \\
				6 & \bf 0.93 & 0.33 & \bf 0.52 & \bf 0.77 & 0.59 & \bf 0.42 & 0.76 \\
				7 & \bf 0.93 & 0.33 & \bf 0.52 & 0.76 & \bf 0.60 & \bf 0.42 & \bf 0.77 \\
				8 & 0.92 & \bf 0.34 & 0.51 & \bf 0.77 & 0.59 & 0.41 & \bf 0.77 \\
				9 & 0.92 & 0.33 & 0.50 & 0.76 & 0.59 & 0.41 & \bf 0.77 \\
				10 & 0.91 & 0.31 & 0.49 & 0.75 & 0.58 & 0.41 & \bf 0.77 \\
				11 & 0.91 & 0.30 & 0.49 & 0.74 & 0.57 & 0.42 & 0.75 \\
				12 & 0.85 & 0.28 & 0.47 & 0.73 & 0.55 & 0.38 & 0.72 \\

				\midrule
				\multicolumn{8}{l}{BART (English); Layers 7--12 are the decoder; std = 0.00 -- 0.01.}\\
				\midrule
				1 & 0.86 & 0.30 & 0.48 & 0.73 & 0.55 & 0.79 & 0.73 \\
				2 & 0.92 & 0.34 & 0.49 & 0.76 & 0.58 & 0.88 & 0.76 \\
				3 & 0.95 & 0.35 & 0.51 & 0.76 & 0.58 & \bf 0.89 & 0.76 \\
				4 & 0.96 & 0.38 & 0.52 & 0.78 & 0.60 & 0.86 & 0.78 \\
				5 & \bf 0.97 & 0.39 & 0.53 & 0.78 & \bf 0.62 & 0.82 & 0.79 \\
				6 & 0.96 & \bf 0.41 & 0.52 & \bf 0.80 & \bf 0.62 & 0.62 & 0.78 \\
				7 & 0.94 & 0.32 & 0.51 & 0.77 & 0.59 & 0.10 & 0.76 \\
				8 & 0.95 & 0.39 & \bf 0.54 & 0.79 & 0.61 & 0.23 & 0.77 \\
				9 & 0.95 & 0.40 & \bf 0.54 & 0.79 & \bf 0.62 & 0.32 & 0.78 \\
				10 & 0.95 & 0.40 & \bf 0.54 & \bf 0.80 & \bf 0.62 & 0.31 & 0.81 \\
				11 & 0.96 & 0.38 & 0.52 & 0.78 & 0.60 & 0.34 & 0.80 \\
				12 & 0.95 & 0.36 & 0.52 & 0.77 & 0.59 & 0.47 & \bf 0.82 \\

				\midrule
				\multicolumn{8}{l}{T5 (English); Layers 7--12 are the decoder; std = 0.00 -- 0.03.} \\
				\midrule
				1 & 0.77 & 0.27 & 0.39 & 0.71 & 0.50 & 0.26 & 0.71 \\
				2 & 0.80 & 0.30 & 0.43 & 0.74 & 0.54 & 0.38 & 0.70 \\
				3 & 0.82 & 0.32 & 0.45 & 0.75 & 0.55 & \bf 0.40 & 0.73 \\
				4 & 0.84 & 0.33 & \bf 0.46 & \bf 0.76 & 0.57 & 0.37 & \bf 0.74 \\
				5 & 0.87 & 0.33 & 0.45 & \bf 0.76 & 0.57 & 0.33 & 0.72 \\
				6 & \bf 0.86 & \bf 0.35 & 0.46 & \bf 0.76 & \bf 0.58 & 0.28 & 0.72 \\
				7 & 0.77 & 0.28 & 0.41 & 0.73 & 0.54 & 0.24 & 0.71 \\
				8 & 0.77 & 0.26 & 0.44 & 0.74 & 0.55 & 0.27 & 0.71 \\
				9 & 0.77 & 0.24 & \bf 0.46 & 0.75 & 0.56 & 0.27 & 0.71 \\
				10 & 0.74 & 0.22 & 0.45 & 0.75 & 0.55 & 0.20 & 0.72 \\
				11 & 0.70 & 0.22 & 0.44 & 0.73 & 0.54 & 0.11 & 0.72 \\
				12 & 0.68 & 0.20 & 0.42 & 0.73 & 0.52 & 0.00 & 0.72 \\
				
				\bottomrule
			\end{tabular}
		\end{adjustbox}
		\captionof{table}{Full results for GPT-2, BART, and T5 over English.}
	\label{tab:result_p2}
	\end{center}

	\vspace{1cm}
	
	\begin{center}
		\begin{adjustbox}{width=0.7\linewidth}
			\begin{tabular}{R{0.7cm}R{1.5cm}R{1.5cm}R{1.5cm}R{1.5cm}R{1.5cm}R{1.5cm}}
				\toprule
				\bf Layer &\bf  NSP & \bf Sent. Ord. & \bf Discourse Conn. & \bf Nuclearity & \bf Relation & \bf EDU segment.  \\
				\midrule
				\multicolumn{7}{l}{Chinese; std = 0.00 -- 0.02.} \\
				\midrule
				1 & 0.30 & 0.38 & 0.36 & 0.53 & 0.58 & 0.70 \\
				2 & 0.47 & 0.38 & 0.36 & 0.53 & 0.58 & 0.75 \\
				3 & 0.47 & 0.42 & 0.36 & 0.53 & 0.58 & 0.79 \\
				4 & 0.83 & 0.42 & 0.36 & 0.53 & 0.58 & 0.79 \\
				5 & 0.90 & 0.44 & 0.35 & 0.59 & 0.60 & 0.81 \\
				6 & 0.93 & 0.44 & 0.39 & 0.60 & 0.64 & \bf 0.83 \\
				7 & 0.94 & \bf 0.45 & 0.42 & \bf 0.64 & 0.67 & \bf 0.83 \\
				8 & 0.94 & 0.44 & 0.43 & \bf 0.64 & 0.66 & \bf 0.83 \\
				9 & 0.94 & 0.43 & \bf 0.46 & 0.63 & \bf 0.68 & \bf 0.83 \\
				10 & 0.96 & 0.43 & 0.44 & \bf 0.64 & \bf 0.68 & \bf 0.83 \\
				11 & 0.96 & 0.42 & \bf 0.46 & 0.61 & 0.66 & 0.81 \\
				12 & \bf 0.98 & 0.40 & 0.44 & 0.62 & 0.66 & 0.78 \\

				\midrule
				\multicolumn{7}{l}{German; std = 0.00 -- 0.07.} \\
				\midrule
				1 & 0.43 & 0.58 & 0.44 & 0.48 & 0.16 & 0.55 \\
				2 & 0.60 & 0.58 & 0.43 & 0.49 & 0.18 & 0.56 \\
				3 & 0.77 & 0.59 & 0.43 & 0.47 & 0.17 & 0.67 \\
				4 & 0.76 & 0.61 & 0.43 & 0.44 & 0.21 & 0.71 \\
				5 & 0.98 & 0.63 & 0.38 & 0.48 & 0.26 & 0.73 \\
				6 & 0.99 & \bf 0.65 & 0.43 & 0.54 & 0.29 & 0.74 \\
				7 & \bf 1.00 & \bf 0.65 & 0.43 & 0.58 & 0.31 & \bf 0.76 \\
				8 & 0.99 & 0.64 & 0.43 & \bf 0.60 & \bf 0.35 & 0.75 \\
				9 & \bf 1.00 & 0.64 & 0.44 & 0.59 & 0.33 & 0.69 \\
				10 & 0.99 & 0.64 & \bf 0.45 & 0.58 & \bf 0.35 & 0.65 \\
				11 & \bf 1.00 & 0.63 & 0.43 & 0.58 & 0.33 & 0.58 \\
				12 & \bf 1.00 & 0.63 & 0.38 & 0.58 & 0.33 & 0.59 \\

				\midrule
				\multicolumn{7}{l}{Spanish; std = 0.00 -- 0.02.} \\
				\midrule
				1 & 0.39 & 0.49 & --- & 0.50 & 0.29 & 0.43 \\
				2 & 0.55 & 0.52 & --- & 0.56 & 0.31 & 0.50 \\
				3 & 0.56 & 0.53 & --- & 0.58 & 0.31 & 0.52 \\
				4 & 0.96 & 0.55 & --- & 0.62 & 0.37 & 0.57 \\
				5 & 0.98 & 0.56 & --- & 0.64 & 0.41 & 0.59 \\
				6 & 0.99 & 0.56 & --- & 0.68 & 0.45 & 0.62 \\
				7 & \bf 1.00 & 0.57 & --- & 0.68 & 0.47 & 0.64 \\
				8 & \bf 1.00 &\bf  0.58 & --- & 0.75 & 0.49 & \bf 0.69 \\
				9 & \bf 1.00 & \bf 0.58 & --- & 0.74 & 0.51 & 0.66 \\
				10 & \bf 1.00 & \bf 0.58 & --- & \bf 0.77 & \bf 0.56 & 0.62 \\
				11 & \bf 1.00 & 0.57 & --- & \bf 0.77 & 0.55 & 0.59 \\
				12 & \bf 1.00 & 0.56 & --- & 0.76 & 0.54 & 0.50 \\
				
				\bottomrule
			\end{tabular}
		\end{adjustbox}
		\captionof{table}{Full results for the BERT monolingual
                  models over Chinese, German, and Spanish.}
		\label{tab:result_other_lang}
	\end{center}

	\clearpage
		
	\begin{center}
		\begin{adjustbox}{width=0.75\linewidth}
			\begin{tabular}{R{0.7cm}R{1.5cm}R{1.5cm}R{1.5cm}R{1.5cm}R{1.5cm}R{1.5cm}R{1.5cm}}
				\toprule
				\bf Layer &\bf  NSP & \bf Sent. Ord. & \bf Discourse Conn. & \bf Nuclearity & \bf Relation & \bf EDU segment. & \bf Cloze ST. \\
				\midrule
				\multicolumn{8}{l}{BERT-Large (English); std = 0.00 -- 0.02.} \\
				\midrule
				1 & 0.34 & 0.26 & 0.29 & 0.60 & 0.37 & 0.23 & 0.61 \\
				2 & 0.53 & 0.25 & 0.33 & 0.62 & 0.40 & 0.28 & 0.67 \\
				3 & 0.57 & 0.26 & 0.32 & 0.63 & 0.42 & 0.29 & 0.66 \\
				4 & 0.60 & 0.29 & 0.32 & 0.63 & 0.42 & 0.40 & 0.66 \\
				5 & 0.64 & 0.30 & 0.35 & 0.65 & 0.43 & 0.43 & 0.69 \\
				6 & 0.82 & 0.31 & 0.37 & 0.66 & 0.44 & 0.45 & 0.68 \\
				7 & 0.87 & 0.31 & 0.39 & 0.66 & 0.45 & 0.44 & 0.69 \\
				8 & 0.95 & 0.30 & 0.39 & 0.66 & 0.45 & 0.44 & 0.72 \\
				9 & 0.96 & 0.32 & 0.40 & 0.67 & 0.47 & 0.43 & 0.70 \\
				10 & 0.96 & 0.32 & 0.40 & 0.67 & 0.46 & 0.49 & 0.71 \\
				11 & 0.97 & 0.32 & 0.39 & 0.68 & 0.47 & 0.53 & 0.70 \\
				12 & 0.97 & 0.33 & 0.47 & 0.77 & 0.57 & 0.58 & 0.71 \\
				13 & 0.97 & 0.34 & 0.52 & 0.78 & 0.61 & 0.56 & 0.71 \\
				14 & 0.97 & 0.34 & 0.57 & 0.81 & 0.65 & \bf 0.59 & 0.73 \\
				15 & 0.97 & 0.34 & 0.61 & 0.82 & 0.67 & 0.58 & 0.75 \\
				16 & 0.97 & 0.35 & 0.60 & \bf 0.83 & 0.67 & 0.55 & 0.75 \\
				17 & 0.97 & 0.34 & 0.62 & 0.82 & \bf 0.68 & 0.53 & 0.82 \\
				18 & 0.98 & 0.36 & \bf 0.63 & 0.82 & \bf 0.68 & 0.54 & 0.82 \\
				19 & \bf 0.99 & \bf 0.37 & \bf 0.63 & 0.82 & 0.67 & 0.50 & 0.83 \\
				20 & \bf 0.99 & 0.34 & \bf 0.63 & 0.81 & 0.67 & 0.48 & \bf 0.84 \\
				21 & \bf 0.99 & 0.35 & \bf 0.63 & 0.81 & 0.65 & 0.41 & 0.83 \\
				22 & \bf 0.99 & 0.35 & 0.61 & 0.81 & 0.65 & 0.37 & \bf 0.84 \\
				23 & \bf 0.99 & 0.34 & 0.59 & 0.80 & 0.63 & 0.36 & 0.82 \\
				24 & \bf 0.99 & 0.33 & 0.57 & 0.77 & 0.58 & 0.31 & 0.81 \\
				
				\bottomrule
			\end{tabular}
		\end{adjustbox}
		\captionof{table}{Full results of English BERT-large.}
		\label{tab:result_large}
	\end{center}
	
\onecolumn
\section{Frozen vs.\ Fine-tuned BERT Layers}
\label{sec:fine-tuned}

\begin{figure*}[ht]
	\centering
	\includegraphics[width=6in]{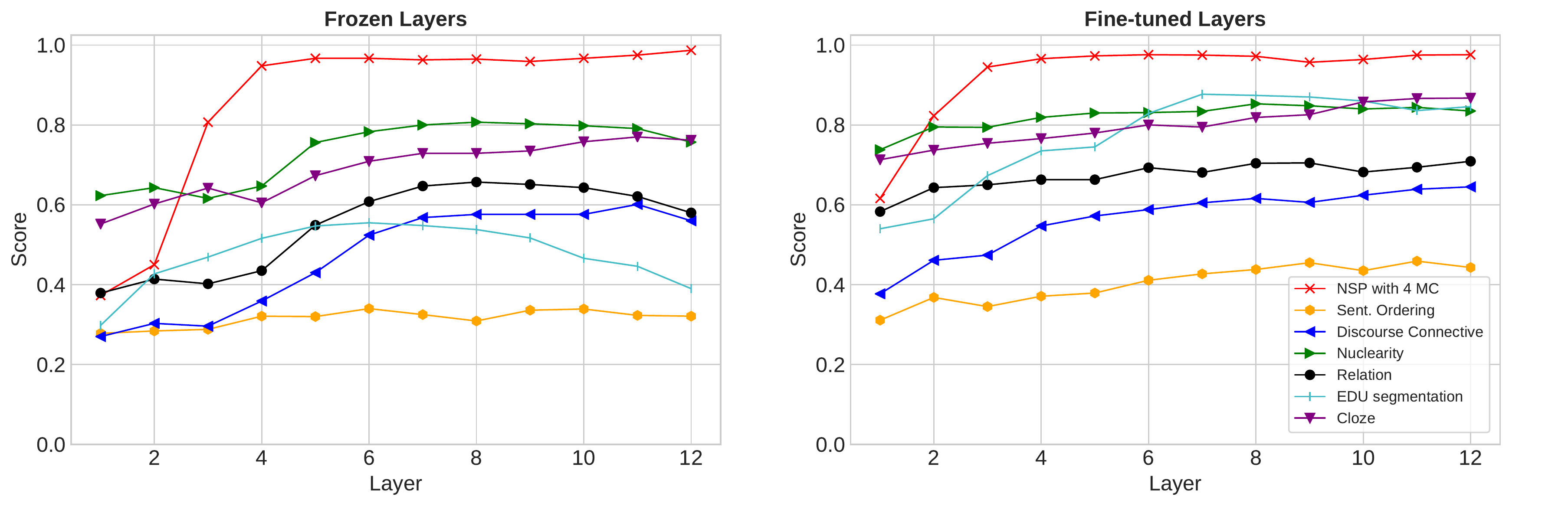}
	\caption{A comparison of BERT with frozen vs.\ fine-tuned layers.}
	\label{fig:finetuned}
\end{figure*}

\section{Full Results of Models with Average Pooling}
\label{sec:cls2}

\begin{figure*}[ht]
	\centering
	\includegraphics[width=6in]{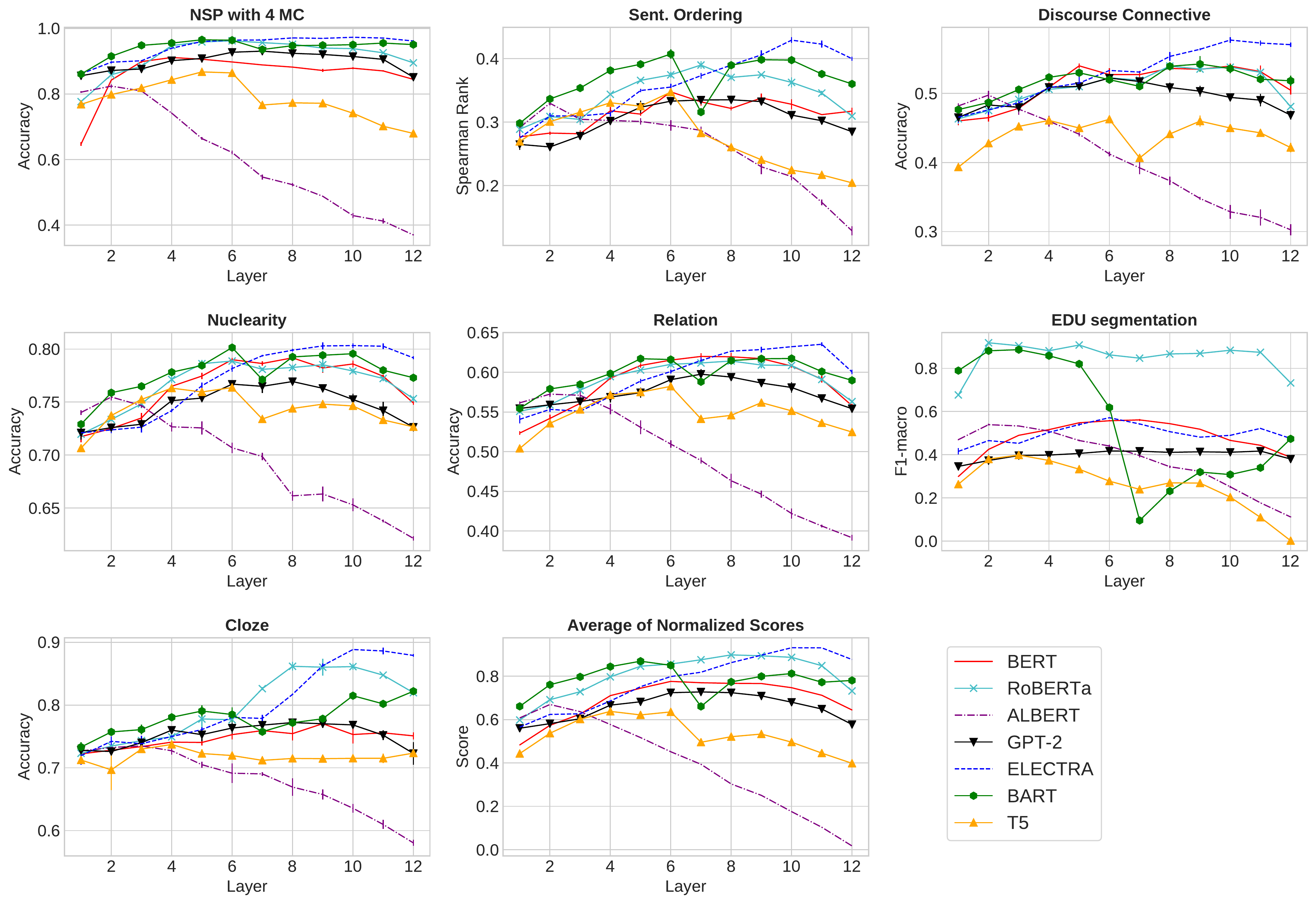}
	\caption{Full results of all models over English with average pooling on all tasks except in EDU segmentation (with the only differences over \figref{result_en}
          being for BERT and ALBERT, where we originally used [CLS] embeddings on two-text classification probing tasks).}
	\label{fig:cls2}
\end{figure*}

\onecolumn
\section{[CLS] vs.\ Average Pooling in English BERT-base Model}
\label{sec:cls}

Average pooling generally performs worse than [CLS] embeddings in the last layers of BERT.

\begin{figure*}[ht]
	\centering
	\includegraphics[width=6in]{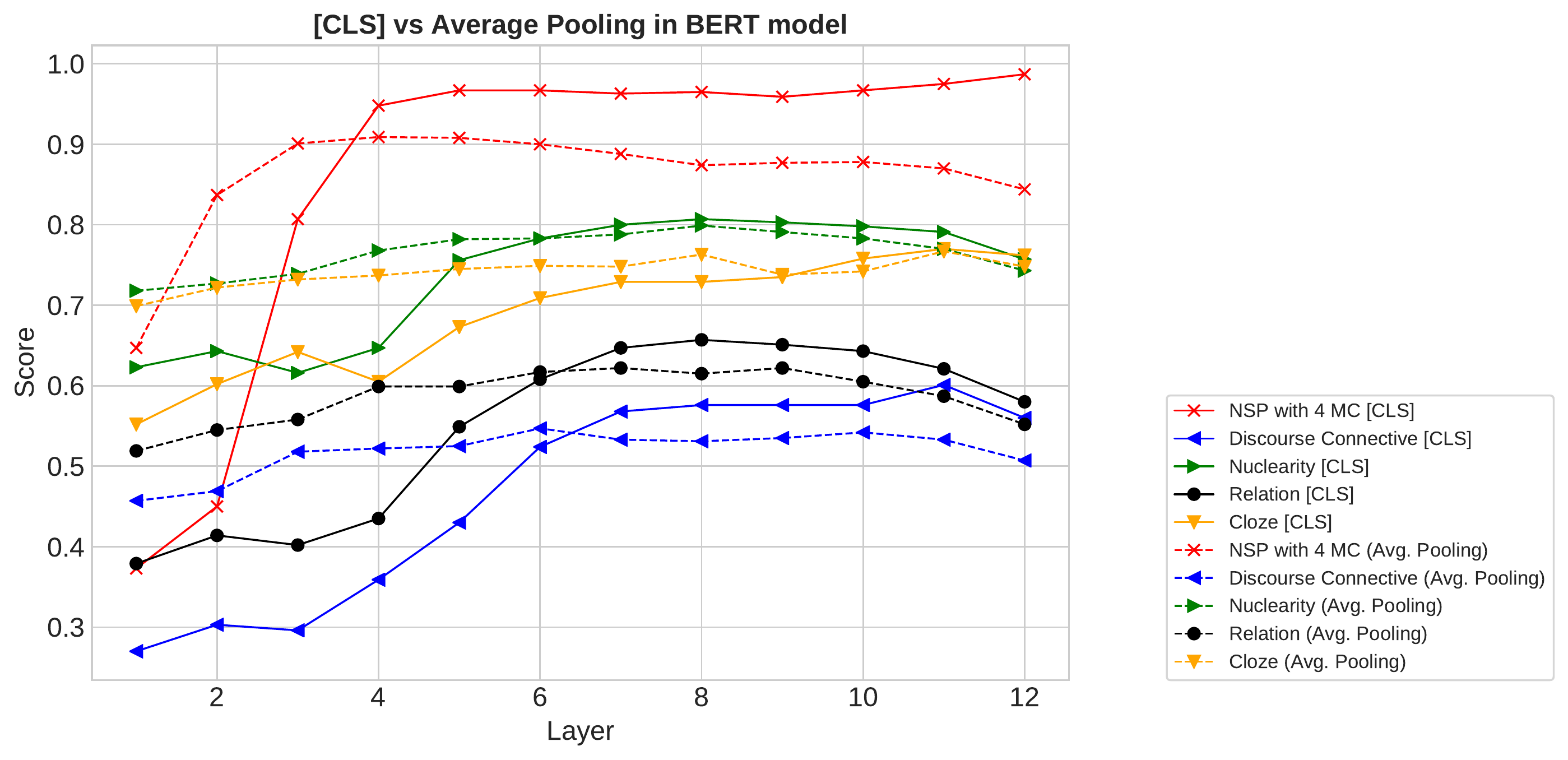}
	\caption{Comparison of [CLS] vs.\ average pooling embeddings for
          BERT-base across the five tasks for English. Please note that sentence ordering and EDU segmentation are always performed with average pooling embeddings and sequence labelling at the (sub)word level, respectively.}
	\label{fig:cls}
\end{figure*}

\end{document}